# Generation of High Spatial Resolution Terrestrial Surface from Low Spatial Resolution Elevation Contour Maps via Hierarchical Computation of Median Elevation Regions

Geetika Barman, B.S Daya Sagar

*[1]Abstract*--We proposed a simple yet effective morphological approach to convert a sparse Digital Elevation Model (DEM) to a dense Digital Elevation Model. The conversion is similar to that of the generation of high-resolution DEM from its low-resolution DEM. The approach involves the generation of median contours to achieve the purpose. It is a sequential step of the I) decomposition of the existing sparse Contour map into the maximum possible Threshold Elevation Region (TERs). II) Computing all possible non-negative and non-weighted Median Elevation Region (MER) hierarchically between the successive TER decomposed from a sparse contour map. III) Computing the gradient of all TER, and MER computed from previous steps would yield the predicted intermediate elevation contour at a higher spatial resolution. We present this approach initially with some self-made synthetic data to show how the contour prediction work and then experiment with the available contour map of Washington, NH to justify its usefulness. This approach considers the geometric information of existing contours and interpolates the elevation contour at a new spatial region of a topographic surface until no elevation contours are necessary to generate. This novel approach is also very low-cost and robust as it uses elevation contours.

*Index Terms*— Digital Elevation Model, Contour Map, Mathematical Morphology, Threshold Elevation Region, Median Elevation Region

## I. INTRODUCTION

It is critical to comprehend the topography of the earth's surface to better understand the changes in shape, texture, or pattern caused by different man-made or natural alterations to the surface[1]. In geoscience, qualitative and quantitative topographic information is widely used because it is an important factor in understanding characteristics such as climate and meteorological characteristics, the spatial distribution of vegetation cover, and various processes occurring at various scales near the surface layer [2]. In summary, a terrain's topography reflects its geological structure [3].

Before digitization, the primary source of data for quantitative terrestrial analysis was topographic maps. Topographic maps were employed to derive geomorphometric quantities, and these quantities were computed via tedious manual methods. Some of the variables that eventually yield geomorphometric quantities include slope, gradient, drainage density, horizontal curvature, and so on. However, as technology advances, people want to display such data in three dimensions, such as ArcView and MapInfo do now, which also provides a richer context of the surrounding topography. So the reconstruction of terrain surface using interpolation becomes a popular study in terrain modeling and analysis. Among other sources, due to their simplicity, contour maps have become a popular way to capture topographic information. The interpolation of the elevation values in the inter-contour space area allows for the reconstruction of a numerical representation of the earth's surface known as Digital Elevation Models (DEMs) [4].

DEMs are two-dimensional discrete functions representing the ground height or elevation of the terrain [5]. Although DEM can now be directly produced from remote sensing data, the generation of terrain surface from contour maps is still popular [6][7]. Firstly, in contrast to satellite data, which might occasionally measure the elevation of natural or man-made objects as ground elevation, contour maps are reliable sources for the true terrain elevation measures [8]. Secondly, topographic maps continue to be a significant source of elevation data for a large part of the earth's surface. Indeed, they are the sole source of knowledge available for historical landscapes. Remote sensing advancement still couldn't reach there or often it is too expensive[9]. Last but not the least, the simplicity and low-cost storage and presentation of larger areas DEM with sparse data like contour is always preferable.

Reconstructing a topographic or terrain surface starts with the basic principle and assumption that it is continuous and smooth [10][11][12]. Thus, most of the reconstruction algorithms interpolate to create a smooth surface from the known existing elevation points [13]. The two most common methods for generating a terrain surface using contour lines are constructing triangulation of points (TIN) on contours or interpolating onto a grid [14]. In this article, our primary focus is on the methods for reconstructing the terrain model from grid-based contour lines which involves interpolating each pixel's elevation value from existing contour pixels [15]. An in-depth discussion of various existing methods for the same is done [16][14][17][18]. However, many of them cannot show satisfactory outcomes in situations such as small ridges, valleys, modeling slopes, etc. Some are prone to over smoothening or

The work of B. S. Daya Sagar was supported by the India-Trento Programme for Advanced Research (ITPAR) - Phase-IV under Grant numbers INT/Italy/ITPAR-IV/Telecommunication/2018.,(*Corresponding author: Geetika Barman.*)

Geetika Barman and B. S. Daya Sagar are with the Systems Science and Informatics Unit, Indian Statistical Institute, Bengaluru 560059, India (e-mail:geetikabarman@gmail.com; bsdsagar@yahoo.co.uk).

some are not able to provide the terrain trends or generate peaks that are not present in the input sample [19].

This article adopts a mathematical morphology-based approach to interpolate a continuous terrain surface by generating intermediate contours from existing contour lines. This method utilizes a non-parametric binary morphological operator and morphological interpolation methods to interpolate the median contours in hierarchical order. In the study [20][21][22], has been established Mathematical Morphology(MM) operators are well suited for visualizing the geometry of terrain surfaces. Although the aforementioned articles did an initial attempt to generate a smooth surface by generating intermediate contours using MM, in this article, we adopted the idea of hierarchically generating a median element set to determine the median contours between the source and target contours. Hausdorff Median Set calculation is conducted along with Threshold Decomposition of the raster form of the source contour map. The major contribution of this article includes the followings: 1) for the first time this binary morphological median set concept is utilized in generating terrain surface from contours using source and target sets extracted from contours which give morphing-like progress of how the source contour changes to the target contour, 2) the idea of Threshold Decomposition makes it lesser expensive in extracting the contour boundary to decide the position of the median contour, 3) simplification of the intermediate contour generation problem to find the median elevation region from the binary set, 4) description of the contour region properties using a spatial and logical relationship in case of different scenarios. This approach of hierarchical recursive generation of intermediate contours to create a smooth and continuous terrain surface also preserves the geometric structure of existing contours and the newly generated intermediate contours. Here we also took care of the flat hilltop and saddle areas whose information are not available in input contours.

## II. BACKGROUND NOTION AND BASIC MM CONCEPT

This section will revisit some fundamental MM operators that are necessary for the suggested framework. For this article, the basic MM operator, dilation, and erosion are only discussed on Binary images. For further theoretical background, readers can refer to [23][24].

### A. Binary Images

A binary image is a set of pixels whose values are either 1 or 0. In a digitization network, the pixels belonging to an object, have a value of 1, and the set of pixels belonging to the background has a value of 0. Formally we can define a binary image as mapping $f: S \to \{0,1\}$ where S is known as the 'Image plane' where an image is defined, also known as 'Definition Domain'. Formally S is a subset of discrete space $\mathbb{Z}^2$ or $\mathbb{R}^2$. All the basic MM operations are introduced here in Binary images only, where these transformations are mapping from the same domain definition as input to integer space i.e. $S \to S$. Instead of point image transformations, MM operations are neighborhood image transformations where the output pixel value is a function of the neighborhood pixel values. The neighborhood is decided by another set of pixels known as the Structuring Element (SE) with a defined origin.

SE is used as a probe to know the morphology of an object. The shape of a SE is generally some basic geometric shape e.g. disk, square, etc., and size (e.g. 3x3 window) is adapted according to the geometry of the object to be processed. In the case of Binary images, the SE used is called 'Flat SE' as they have a set of pixels without any valuation. Given a set $E \subset \mathbb{R}^2$, a flat SE is defined as

$$f(x) = \begin{cases} 0, & x \in E \\ -\infty, & x \notin E \end{cases} \quad (1)$$

Throughout this article, we shall only consider SE as flat SE and disk of a radius of 1.

1) *Dilation*: Answering the question 'Does the SE hit the set' gives us the definition of dilated set. The *Dilation* of set $Y$ by SE $B$, $\delta_B(Y)$ can be defined as:

$$\delta_B(Y) = Y \oplus B = \{y | B_y \cap Y = \phi\} = \bigcup_{-b \in B} Y_b \quad (2)$$

$Y_b$ is the translation of $Y$ along the vector $b$

2) *Erosion*: Similar to *dilation*, another question 'whether the *SE* fits the set or not gives us the *Eroded Set*'. Thus *Erosion* of set $Y$ by SE $B$ can be defined as:

$$\mathcal{E}_B(Y) = Y \ominus \hat{B} = \{y | B_y \subseteq Y\} = \bigcap_{b \in \hat{B}} Y_{-b} \quad (3)$$

In (2) and (3), $Y \oplus B$ is *Minkowski addition* and $Y \ominus \hat{B}$ is *Minkowski subtraction*. If *SE* is symmetric to its origin, then both *erosion* and *dilation* are equivalent to *Minkowski subtraction* and *Minkowski addition* respectively.

The SE mentioned here is $\lambda B$, where $\lambda$ is the size of the structuring element and $\lambda = 0,1,2,...,N$. Equations (2) and (3) can be explained also in their multiscale version as a sequence of repetitive operation with smaller SE as mentioned in (4). For example, *dilation* and *erosion* of set $Y$ with a SE of size n are equivalent to *dilation/ erosion* of set $Y$ n-times with the same SE of size 1. Formally we can write it as:

$$\delta_n(B) = \delta_B(n) \quad (4)$$

### B. Hausdorff Distance

In Mathematical morphology, the distance between two sets $X$ and $Y$ can be defined in terms of classical Hausdorff distance as $max\{\sup_{x \in X} d(x,Y); \sup_{y \in Y} d(y,X)\}$. As mentioned in [25], if two sets $X$ and $Y$ are non-empty and compact, then Hausdorff distance also can be written as dilation and erosion by a SE of size $\lambda$. Corresponding Hausdorff distances are known as *Hausdorff dilation distance* $(\rho(X,Y))$ and *Hausdorff erosion distance* $(\sigma(X,Y))$. Algebraically these two distances are dual to each Other and follow all the properties of a distance metric

$$\rho(X,Y) = \inf\{\lambda: Y \subseteq \delta_\lambda(X); \; X \subseteq \delta_\lambda(Y)\} \quad (5)$$

$$\sigma(X,Y) = \inf\{\lambda: \epsilon_\lambda(Y) \subseteq X; \; \epsilon_\lambda(X) \subseteq Y\} \quad (6)$$

## C. Median Set Computation

Jean Serra introduced a method to compute the median set in [25] using the *Hausdorff dilation distance ($\rho$)* and *Hausdorff erosion distance($\sigma$)* in (5) and (6). It is also known as Serra's median. Given two non-empty and ordered nested sets, $X$ and $Y$ such that $X$ is completely contained in $Y$ i.e. $X \subseteq Y$, the median between these two sets can be computed as:

$$M(X,Y) = \cup\{(X \oplus \lambda B) \cap (Y \ominus \lambda B), \quad \forall \lambda \geq 0\} \quad (7)$$

During the computation of the median set, the concept of multiscale erosion and dilation is used. We can say that M is midway from X and Y as for every point $m \in M$, $\exists \lambda \geq 0$ such that $d(m, X) \leq \lambda$ and $d(m, Y^c) \geq \lambda$. From this observation, it can be implied that this median set follows a symmetrical property.

*Property 1:*

The median set $M(X, Y)$ is at $\mu$ Hausdorff dilation distance from $X$ and $\mu$ Hausdorff erosion distance from $Y$. $\mu$ is defined as:
$$\mu = \inf\{\lambda: \lambda \geq 0, (X \oplus \lambda B) \supseteq (Y \ominus \lambda B)\}$$

From the above-mentioned property, it can be claimed that, if X and y are ordered and non-empty set and $X \subseteq Y$, then also $X \subseteq M(X,Y) \subseteq Y$.

## D. Threshold Decomposition

A grey-scale image signal $f(i,j)$ is assumed to be a non-negative 2D sequence, with intensity values, $q = 0,1,2,\ldots,Q$, then $Q + 1$ threshold binary images are obtained by f $f$ at all possible intensity values $0 < q < Q$ such that,

$$f_q(i,j) = \begin{cases} 1, & f(i,j) \geq q \\ 0, & f(i,j) < q \end{cases} \quad (8)$$

The reconstruction of $f$ can also be done from the threshold binary image,

$$f(i,j) = \sum_{q=1}^{Q} f_q(i,j) \; \forall i,j = \max\{q: f_q = 1\} \quad (9)$$

So, applying any image transformation $\beta$ to the threshold image and original image $f$ gives [26]:

$$\beta[t_q(f)] = t_q[\beta(f)] \; where \; t_q(f) = f_q \quad (10)$$

## III. METHODOLOGY

The problem of surface generation from a contour map can be formally stated as given a pair of contours as $C_i$ and $C_j$ with their respective elevation values as $\mathcal{E}_i$ and $\mathcal{E}_j$ such that $\mathcal{E}_i < \mathcal{E}_j$ and defining an inter-contour space as $\Phi$, then the aim is to generate all possible intermediate contours $C_k \in \Phi$, and to reconstruct a surface $\psi$ over the image such that $\forall p \in \Phi$, $\mathcal{E}(p) = \mathcal{E}_i \; if \; p \in C_i$ and $\mathcal{E}(p) = \mathcal{E}_j \; if \; p \in C_j$ and $\mathcal{E}(p) = [\mathcal{E}_i, \mathcal{E}_j] \; if \; p \in \Phi$ and $\mathcal{E}_j < \mathcal{E}_k < \mathcal{E}_j$. This statement can be extended to $n$ numbers of contours such that $C_i, i \in n \; and \; n > 2$.

Contour morphology and contour line properties play a vital role in producing intermediate contours between any successive contours. The proposed approach is also based on some given contour properties and assumptions that are to be discussed in this section. Further, a series of simplifications of the problem of intermediate contours generation to median set computation is also to be discussed here.

### A. Contour lines

Contour lines or Isolines are connected points having the same elevation value throughout the connected points. We assume that all the contours in the contour map are well-sampled and well-connected segments. Contour lines possess some inherent properties. Some of the contour properties include:

1) Two contours of different elevation values never intersect each other. Generally, the contours approximately run parallel [27].
2) Contour lines are always increasing or decreasing corresponding to their elevation value and it has an ordered spatial relationship also. It always follows the child-parent or inside-outside relationship, e.g. if $C_i \; and \; C_j$ are two contours and if their spatial relationship can be described as $C_i \; is \; inside \; C_j$ or $C_j \; is \; outside \; C_i$, then it is always maintained an ordered relationship for their elevation values, $e(C_i) \geq e(C_j) \; or \; e(C_j) \geq e(C_i)$ [28].
3) *Contour Region* – An area surrounded by a closed contour is known as the *Contour Region* [28]. Every contour is closed even if it is not within the region of the display, then at least outside of the displayed window. So we assume the boundary of the raster window is the close boundary of the contour line if it is not closed within the window of the display.
4) Any point *p* in the neighborhood of a contour point can have only two values, either greater than or equal to the contour elevation value or less than the contour elevation value. If all the points in a contour region are assumed to have the same elevation value, then it means that the region is assumed to be flat.

All these aforementioned properties carry important information for generating Median Elevation Region (MERs) and subsequently in the generation of intermediate contours.

## IV. INTERMEDIATE CONTOUR GENERATION

Based on the properties of the contour map discussed in section III, the objective mentioned in this article can be achieved by the sequential steps as follows:

5) Convert the given contour map from vector format to raster, where contour lines are represented in terms of the grid.

6) Extraction of binary images, Threshold Elevation regions (TERs) based on their elevation values of the contour lines.
7) Establishing a spatial relationship between *TERs* extracted and their categorization.
8) Computation of Median Elevation Region (MERs) hierarchically using the Median Set computation method.
9) Generation of a sequence intermediate contours based on the generated MERs for every pair of input TERs.

A complete algorithm to generate the maximum possible intermediate contours is as follows:

---
**ALGORITHM1: COMPUTE INTERMEDIATE CONTOURS**

**Input**: Raster Image with a set of contours $C_i$
**Output**: Grid of Interpolated heights/ digital elevation model
1. $\mathcal{W} \leftarrow$ Input image
2. $e_i \leftarrow$ elevation of contour $C_i$
3. $e_j \leftarrow$ elevation of contour $C_j$
4. **Elv_Reg** $\leftarrow$ total number of contours in $\mathcal{W}$
5. **for all $i$ in Elv_Reg:**
6.     Initialize empty queue $Q$
7.     $S \leftarrow$ Intercontour Space between $C_i$ and $C_{i+1}$
8.     $\mathcal{T}_i \leftarrow$ compute TER for contour $i$
9.     $\mathcal{T}_{i+1} \leftarrow$ compute TER for contour $i+1$
10.     $Q.push((\mathcal{T}_i, \mathcal{T}_{i+1}))$
11.     **while** $(e(S).any() < 0)$: //$e(S)$ is all pixel value in $S$
12.         $\mathcal{P} \leftarrow Q.pop(0)$
13.         $\mathcal{T}_i \leftarrow \mathcal{P}(0), \mathcal{T}_{i+1} \leftarrow \mathcal{P}(1)$
14.         $\mathcal{M} \leftarrow$ compute MER for $(\mathcal{T}_i$ and $\mathcal{T}_{i+1}$ )
15.         $C_m \leftarrow$ morphological gradient of $\mathcal{M}$
16.         $e_m \leftarrow 1/2(e_i + e_j)$
17.         map $C_m$ to $\mathcal{W}$
18.         $Q.push((\mathcal{T}_i, \mathcal{M}), (\mathcal{M}, \mathcal{T}_{i+1}))$
19.     **end** while
20. **end** for

---

*A. Threshold Elevation Regions and their extraction*

Our objective in this article is to generate a surface of higher spatial resolution from the available sparse contours, in other words, low spatial resolution elevation contours. From the properties of isolines or contour lines, it ensures that two contour lines are always approximately parallel which creates a constant contour interval. Moreover, the elevations rarely deviate from the slope defined by the contours [28]. As mentioned in earlier literature the surface can be generated by inserting new contours midway between the successive contours. Our proposed algorithm computes these new intermediate contours by computing hierarchically the sequence of the Median Elevation Regions (MERs) using median set computation between two successive contour lines. Since the new intermediate contour we generate is approximately midway between two successive contours, the new intermediate contour is assigned an elevation of the mean of the considered two successive contours.

The methodology mentioned here is applied to grid data of contour lines. All the topographic maps are available as a shape file. In the very first step, all the shape files are converted to raster data. The set of contour lines is then presented as a Grid of data.

The raster image obtained from processing the topographic map can be considered a grayscale image with intensity values of the elevation of the contour lines. Instead of processing the grid of the contour map as a grayscale image, a procedure is proposed here for generating binary Input sets of contour regions from a limited set of layers i.e. the input set of contours.

The contour information presented in a given contour map can be considered as spatial layered information of an object or event. This information can be ordered, semi-ordered, or disordered. For example, if we have two contours $C_i$ and $C_j$ in our input contour map, the input set information extracted from these two is denoted by $X_i$ and $X_j$ as respectively, known as *source set* and *target set*. Consequently, all the subsets present in the set $X_i$ and $X_j$ are denoted as $X_i^1, X_i^2, \ldots, X_i^3$ and $X_j^1, X_i^2, X_i^3, \forall i \in N$.

A simple method is suggested here for extracting the input sets from the input contour map. Based on (8) to (10), we generate TERs for each available elevation value. Let us consider the input contour map as a 2-D sequence $f(x,y)$ which assumes all the elevation levels as intensity values: $I = e(C_i), i = 0,1,2,\ldots n$. At all possible intensity levels of $I$, we threshold the image $f(x,y)$ such that we obtain the set of threshold binary images

For all $C_i, i = 0,1,\ldots,n$

$$f_I(x,y) = \begin{cases} 1, & f(x,y) = I \\ 0 & f(x,y) \neq I \end{cases} \quad (11)$$

The set of binary images obtained using (11) gives us all the contours at each elevation level. We convert them to threshold elevation regions (TERs) by considering the regions under the contours as flat regions i.e. with the same elevation value.
Let us assume $C_i$ and $C_{i+1}$ are two consecutive contours and $e(C_i)$ and $e(C_i)$ are their respective elevation values
Case 1: $e(C_i) > e(C_{i+1})$:

$$TER_i = \mathcal{T}_i = f_I(m,n) = \begin{cases} 1, if\ f(m,n) \leq e(C_i) \\ 0, otherwise \end{cases} \quad (12)$$

Case 2: $e(C_i) < e(C_{i+1})$:

$$TER_i = \mathcal{T}_i = f_I(m,n) = \begin{cases} 1, if\ f(m,n) \geq e(C_i) \\ 0, otherwise \end{cases} \quad (13)$$

The obtained $TER, \mathcal{T}_i$ follows some property that either $T_i \subset T_{i+1}$ or $T_{i+1} \subset T_i$. Here, we reduce the problem of finding intermediate contours between two successive contours to an easier form, instead of interpolating from a contour $C_i$ to its next immediate contour $C_j$, we now interpolate between two successive TERs ($T_i$ and $T_{i+1}$). Fig. 1. (a), (d), (g) shows three different examples of possible nested contours, (b), (c) are two TERs obtained from (a), similarly (e), (f) are obtained from (d), and (h) and (i) are corresponding TERs of (g). All TERs extracted are nothing but binary images. And we further can assume binary images as set *X* and *Y* as discussed earlier.

## C. Spatial Relationship between TER

Let $T_1, T_2 \ldots, T_n$ be the TERs corresponding to the contours

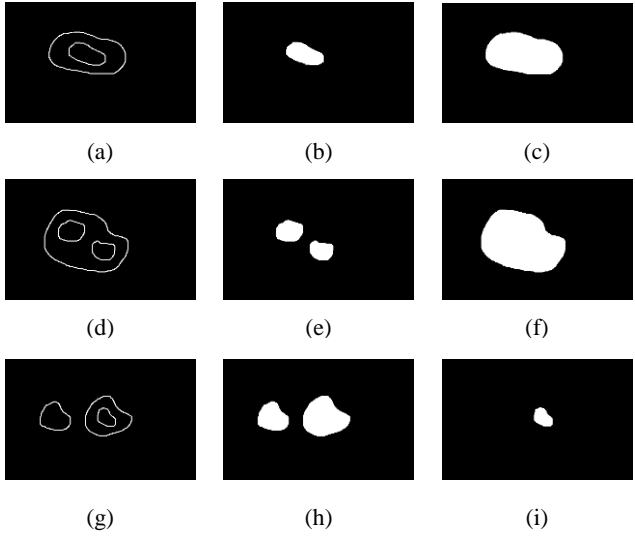

Fig. 1. (a), (d), (g) represent different possible cases of nested contours; (b), (c) represents corresponding TERs of (a); (e), (f) represents that of (d) and (h),(i) represents that of (g).

$C_1, C_2, \ldots, C_n$. All the TERs obtained nothing but the binary images, as we can recall. These TERs are represented as sets. If $T_i \subseteq T_j$ or $T_j \subseteq T_i$, then $T_i$ and $T_j$ are ordered sets. Also the same applied to the subset $(T_i^1, T_i^2, \ldots, T_i^n)$ or $(T_j^1, T_j^2, \ldots, T_j^n)$ embedded within each set. If $T_i$ and $T_j$ are partially contained in each other, then it is semi-ordered. If there is no intersection between $T_i$ and $T_j$, then disorder it is.

Intuitively, the sets $T_i, i = 0, 1, \ldots, n$ represent the elevation profile of the cross-section of the area at a specific spatial position $i$. The profile evolves based on its slope to the next elevation profile at another spatial position $j$ where $j = i + 1$. With this assumption, we can study every possible spatial relationship between consecutive TERs $T_i$ and $T_j$; $j: i + 1$. Based on the different spatial relationships, TER can be grouped into some categories as follows:

1. This category satisfies simple conditions of a spatial relationship between $T_i$ and $T_j$, where $T_i \subseteq T_j$ or $T_j \subseteq T_i$ and $T_i^n \cap T_j^m \neq \phi \ \forall n, m$. this category includes the criteria for its corresponding subsets/ connected components, a)$\forall n, m, T_i^n \subseteq T_j^m$ or $T_j^m \subseteq$, where $n = m$. b) $\forall n, m, T_i^n \subseteq T_j^m$ or $T_j^m \subseteq T_i^n$, but $n \neq m$. The criteria mentioned in b) can arise is shown in an example Fig. 1. (d) and (g).
2. In some nested cases, where $n \neq m$ and there exist some n or m such that, $T_i^n \cap T_j^m = \phi$. An example of such a case is the hilltop area as shown in Fig. 1. (h), where the left-side TER is a probable case of a hilltop or the ultimate TER consists of the hilltop contoured area or the saddle point area which does not have the corresponding TER to interpolate with.

Above mentioned category-1 globally belong to the set category $T_i \cap T_j \neq \phi$, whereas category 2 belongs to $T_i \cap T_j = \phi$ assuming that some of the subsets or connecting components are an empty set.

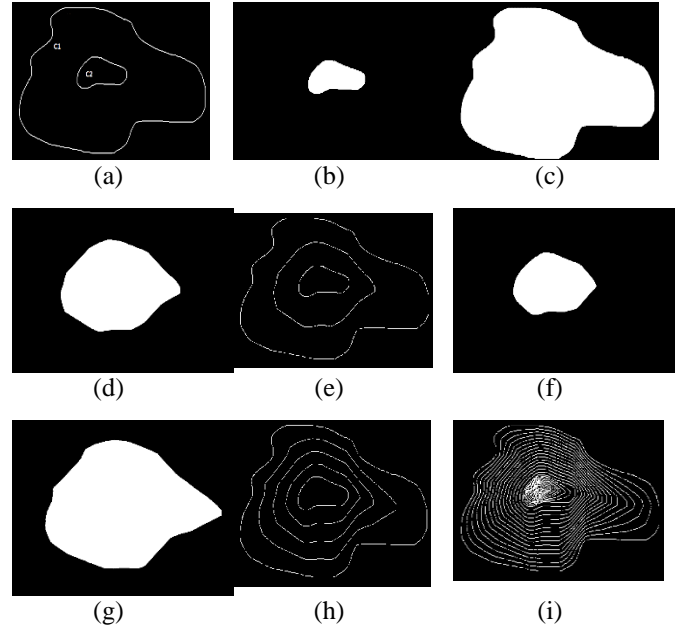

Fig. 2. (a) a pair of contours $C_1$ and $C_2, e(C_1) < e(C_2)$ where $e(C_i)$ is elevation of contour $C_i$; (b) and (c) are corresponding TERs of (a); (d) MER(M) obtained after 1st iteration by using the proposed method; (e) Intermediate contour obtained using morphological gradient on MER; (f) and (g) are 2nd level MERs between $C_1, M$ and $M, C_2$; (h) contours generated after 2nd iteration and; (i) Intermediate contours generated after 3rd iterations;

### D. Intermediate Contour Computation using Median Set

Serra's median mentioned in (7) considers the input sets $X$ and $Y$ globally. But, as we can see from spatial relationships between TER, the interpolation of subset layers from the layered input contour sets are category dependent. To better visualize the spatial transition of a subset from one spatial location to another. We compute the median layers between successive layers (TER), and we use (7) category-wise. The generation of MER and its corresponding intermediate contours on synthetic data are shown in Fig. 2

Case 1:
  a. $T_i \subseteq T_{i+1}$

$$MER(T_i, T_{i+1}) = \bigcup_\lambda ((T_i^n \oplus \lambda B) \cap (T_{i+1}^m \ominus \lambda B)) \quad (14)$$

  b. $T_{i+1} \subseteq T_i$

$$MER(T_{i+1}, T_i) = \bigcup_\lambda ((T_i^n \ominus \lambda B) \cap (T_{i+1}^m \oplus \lambda B)) \quad (15)$$

Case 2:
  Under unique situations as explained in category 2, the MER can be computed as:

$$MER(T_{i+1}, T_i) = \bigcup_{\lambda}^{N}((UT_{i+1}^n \ominus \lambda B) \cap (T_{i+1}^m \oplus \lambda B)) \quad (16)$$

Where $UT_{i+1}^n$ is the ultimate eroded version of $T_{i+1}$.

For interpolating the hill-top regions, the TERs obtained using (16), is used as the TERs of the next contour, and the elevation of that region is computed as the previous elevation added to the contour interval.

In the case of category 2, where there is no successive TERs or corresponding subsets, to compute the intermediate contours, the MERs is calculated by adding the ultimate erosion of the TER ($T_{i+1}$) or its any subset ($T_{i+1}^n$) whose corresponding successive subset is empty.

Finally, the intermediate contours are computed from the MERs obtained from (14)-(16), using the Morphological Gradient (MG). The intermediate contour of two contours $C_i$ and $C_{i+1}$ is computed as:

$$\mathcal{IC}(C_i, C_{(i+1)}) = MG(MER(C_i, C_{i+1})) \quad (17)$$
$$= (MER \oplus B) - (MER \ominus B)$$

where B is the structuring element

*F. Intermediate contour and Time complexity*

Eq (11)-(17) clearly shows the computation of MER and subsequently the intermediate contour computation from its successive input TERs. Let us assume that, the first input contours $C_i$ and $C_j$ are considered as $C_o$ and $C_1$ respectively, so the first intermediate contour can be considered as $C_{0.5} = \mathcal{IC}(C_0, C_1)$. The $C_{0.5}$ is midway between $C_1$ and $C_0$ and it breaks down the intercontour space into half. Next, we can compute the intermediate contours between $C_{0.5}, C_1$; and $C_0, C_{0.5}$ which are respectively midway between $C_0, C_{0.5}$ and $C_{0.5}, C_1$ Thus, the recursive computation of the intermediate contour divides the contoured space into half of its original width. If $N$ is the contour space then to compute all the intermediate contours, it takes $\log_2 N$.

IV. RESULTS AND DISCUSSIONS

The proposed method mentioned in Section III is applied to real contour data to demonstrate its applicability. An overview of how the method works in synthetic data and its result is already shown in Fig.2. Fig. 2(a) is given input contours $C_1$ and $C_2$, 2(b) and 2(c) are respective TER, 2(d) is first level MER between 2(b) and (c) using proposed method, 2(f),2(g) are corresponding second level MER between 2(b),2(d) and 2(c),2(d), 2(e),2(h) are the corresponding contours from their MER and 2(i) is intermediate contours after 3rd level of intermediate contour generation. This method also interpolates intermediate contours of hilltop regions using (16) as shown in Fig. 2(i). For the best applicability, we consider some sparse contour data and examine how they can produce dense spatial distribution or a continuous grid of interpolated elevation values. We consider demonstrating the proposed method to a set of contours extracted from a portion of the topographic map from the USGS DLG [29]. The procedure is applied to Grid data only. Since the proposed method is working only on grid data, at first the obtained contours are preprocessed to raster format using QGIS and ArcGIS software. During the conversion of vector to a raster format, there might be some errors, which is inevitable. Because of these errors, some undesired artifacts and errors may have been formed on occasion.

*A. Case study on Mt. Washington NH*

Two sets of contours named "Zone A" (800x800 grid, Fig. 3(a)) and "Zone B" (500x500 grid, Fig.3(d),) from the topographic map of Mt. Washington [29], are chosen to demonstrate the ideas mentioned above in section III. The contour interval is 20 meter in both cases and the elevations are ranging from 2300-3540 meter in Zone A and 1140-1920 meter in Zone B respectively. In both category 1 and category 2, the spatial relation of TER can be observed in Zone A. whereas, in Zone B, the TER obtained, can be classified as category 1 only. The topographic surface is computed in both zones by computing the maximum possible intermediate contours between successive TERs using (11) – (17). Fig.3 (b) and Fig.3 (e) are the topographic surfaces computed using the proposed method from Figs. 3(a) and 3(d) respectively. All the necessary codes and results are available at[30]. For better visualization, 3D- rendering of the obtained surfaces using QGIS software[31] is also shown in Figs. 3(c) and 3(f). It can be seen that the topographic surface generated using the method is rather smooth and without any contour ghosting or artifacts.

*B. Validation of the Intermediate contours*

Since the available contour maps do not have any ground truth data to validate with, for validation of the quality of generated intermediate contours, we created some test instances by skipping random alternate contours from original contours (Figs. 3(a),3(d)), resulting in a contour map with 40-meter contour intervals in some areas. The test instances where we have the ground truth value comprise the skipped set of contours. Then, to assess the quality of the interpolated contours using the proposed method, we compare the available ground truth value of the set of contours with the interpolated contours at the test points using some of the error measures (Fig. 4). To check the quality of the proposed method, we compute $\mathcal{IC}(C_i, C_{i+2})$ (interpolated contours) with ground truth $C_{i+1}$ for all possible $i$ values of contours. If the comparison yields an acceptable degree of match, we can conclude that the generation of further levels of intermediate contours will undoubtedly provide a better quality surface from available contours. To compare $C_{i+1}$ with $C_{i+1} = \mathcal{IC}(C_i, C_{i+2})$, we generated test cases for Zone A (Fig. 3(g)) and Zone B (Fig. 3(j) by skipping alternate $C_i$ randomly resulting in 40-meter contour interval.

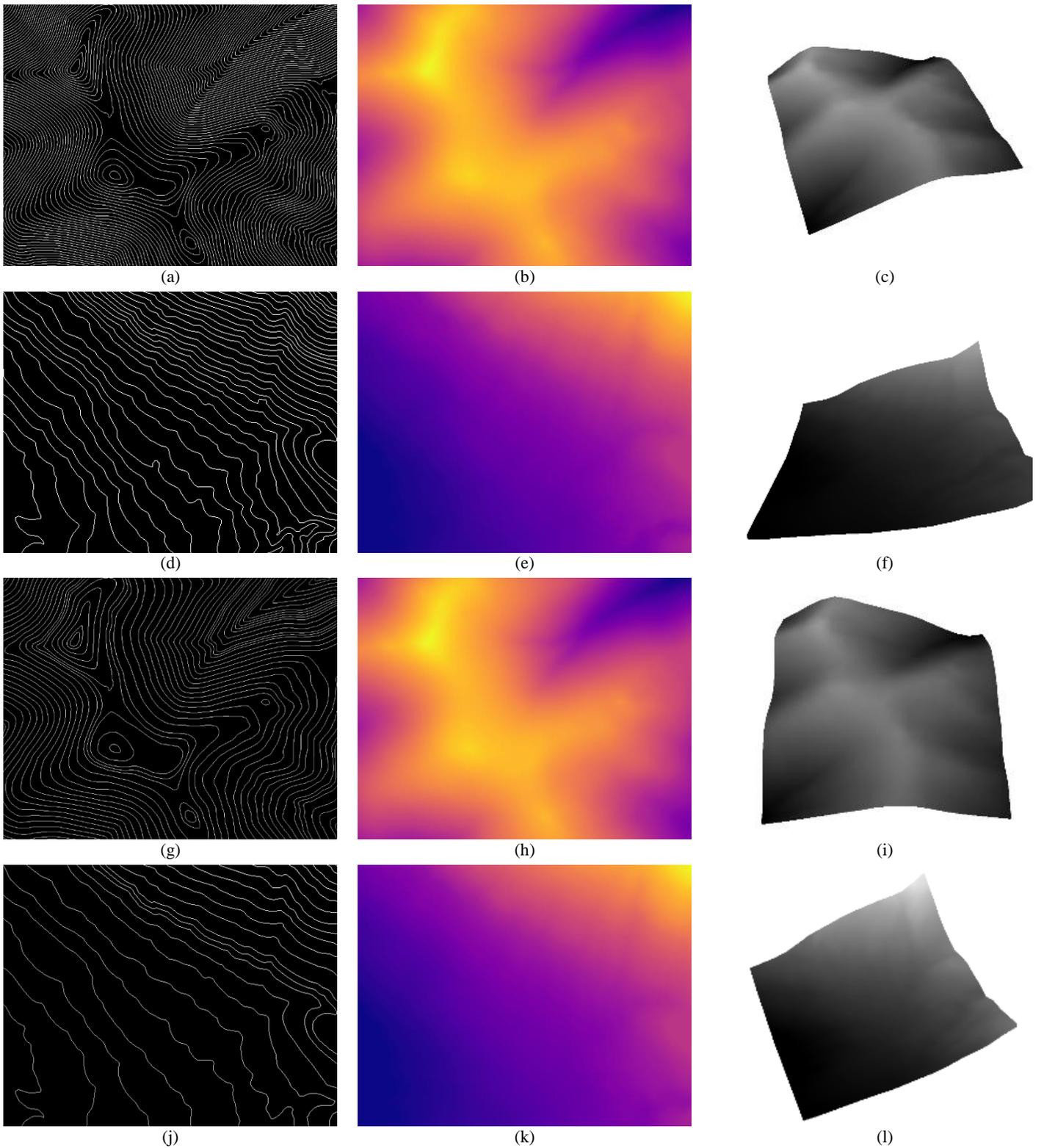

Fig. 3. (a) Given input set of contours "Zone A"; (b) computed topographic surface from (a) using the proposed method; (c) 3D rendering of the surface in (b); (d) Given input set of contours "Zone B"; (e) Computed topographic surface from (b) using proposed method; (f) 3D rendering of the surface obtained in (d); (g) the set of test contours of contour interval 40 meter obtained from (a) by skipping alternate contours $C_i$ from input contours; (h) and (i) are corresponding computed topographic surface and 3D rendering of the surface respectively using the proposed method; (j) the set of test contours by skipping alternate contours $Ci$ from input contours "Zone B" (b); (k),(l) are respective interpolated topographic surface and it's 3D view from (j).

From each of the test sets, by comparing the given $C_{i+1}$ and $\mathcal{IC}(C_i, C_{i+2})$, using the proposed method, we discovered a significant resemblance for the interpolated intermediate contours. Fig. 5(a) and 5(b) depicts a portion of the test set of contours that illustrates the visual representation of interpolated 1st-level intermediate contours by superimposing the computed intermediate contours ($\mathcal{IC}(C_i, C_{i+2})$, blue in color with the ground truth contour $C_{i+1}$, red in color. From the superimposition of computed and original contours (Fig. 5(b)), it can be observed that there is a small difference between the actual and the interpolated contours because of the changes in the shape of the contours and absence of the corresponding contours in the test case. From the visualization, one can assume that the proposed method interpolated a quality of intermediate contours. Further, for sake of better comparisons, the Root Mean Square Error (RMSE) and Mean Absolute Percentage Error (MAPE) were computed for the surface generated from the test set of contours (Fig. 3(g), 3(j)) for both "Zone A" and "Zone B" with the surface generated (Fig. 3(b),3(e)) from given input set of contours. The RMSE and MAPE computed the accuracy of the interpolated intermediated contour values at the set of test contours (25 contours for "Zone A" and 18 contours for "Zone B" as test case) for every set of $\mathcal{IC}(C_i, C_{i+2})$ and $C_{i+1}$ (Fig. 4). The interpolated elevation values that intersects the original contours must have the values equal to or approximately equal to the contour labels, when measured by RMSE and MAPE. The lesser the RMSE and MAPE value between actual contour and interpolated contour, the more they are similar and the more valid the interpolation is. The computed RMSE (Fig. 4) for all the test contours $C_{i+1}$ for all $i$ lie between 2 to7 for "Zone A" and 1to 6 for "Zone B" and the maximum MAPE obtained for "Zone A" and "Zone B" are 0.152 and 0.326 respectively. Also, the minimum MAPE is 0.039 and 0.072 respectively for "Zone A" and "Zone B" (Fig.4. for RMSE and MAPE for both cases). The maximum value for both RMSE and MAPE is reasonably acceptable. We also have analyzed that, for "Zone A" more than 80% of the interpolated contours have RMSE values from 2- 4 and MAPE less than 0.1% which is definitely within an acceptable range. Similarly, for "Zone B", 80% of interpolated contours have RMSE values less than 5 and 90% have MAPE values less than 0.3% as described in Fig. 4. It indicates that the interpolated set of test contours ($\mathcal{IC}(C_i, C_{i+2})$) matches the original set of contours to a high degree and the error of the proposed interpolation method is within an acceptable range.

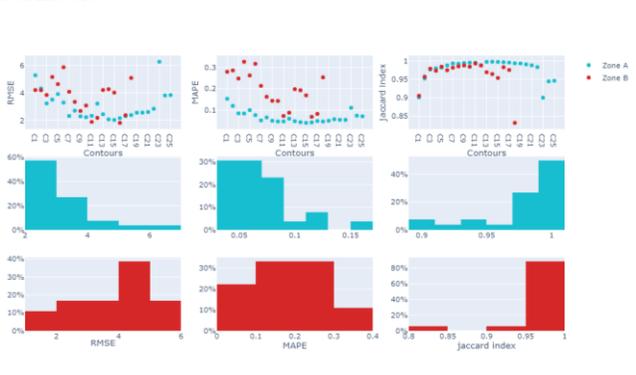

Fig. 4. The figure depicts the RMSE, MAPE, and Jaccard Index value for each test contour for both case study "Zone A" and "Zone B"; 2nd and 3r row plots the percentage-wise distribution of interpolated test contours for RMSE, MAPE, and Jaccard Index values of "Zone A" and "Zone B" respectively, red color represents "Zone A" and blue represents "Zone B".

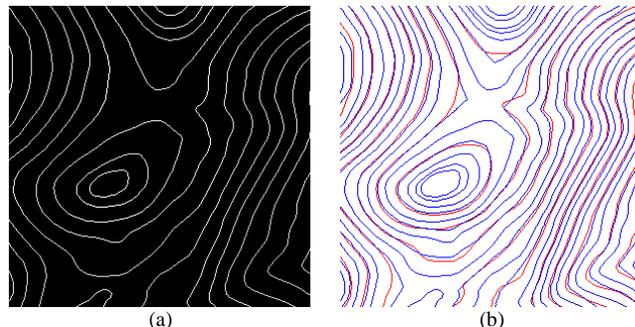

(a)                      (b)

Fig. 5. (a) Set of test contours of contour interval 40 meter obtained from Zone A; (b) Overlapping of interpolated 1st level intermediate contours (blue color) and original contours (red color).

This method computes all feasible intermediate contours between two given contours, so the geometric shape of the contours is also essential. Therefore, to validate the quality of the computed intermediate contours, Hausdorff Distance and

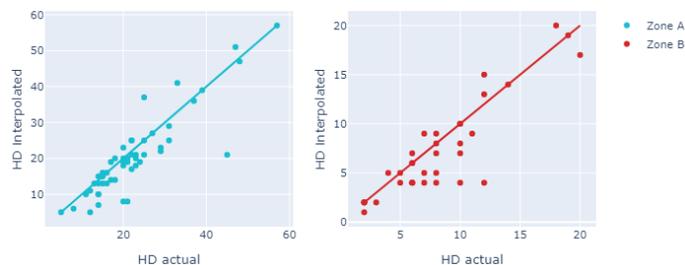

Fig. 6. The scatter plot of Hausdorff Distance values for $\mathcal{IC}(C_i, C_{i+2})$ and $C_{i+1}$ (as HD interpolated) vs. Hausdorff Distance values for $(C_i, C_{i+1})$ (as HD actual) for all the contours, blue points indicating Zone A contours and red point indicate Zone B contours

Jaccard Index were also computed between $\mathcal{IC}(C_i, C_{i+2})$ and $C_{i+1}$ for all $i$ (Table. I, Fig. 4, Fig.6). The lower the difference between the values of Hausdorff Distance (HD) of ($\mathcal{IC}(C_i, C_{i+2})$ and $C_{i+1}$) and Hausdorff Distance (HD) of $(C_i, C_{i+1})$, and approximate to 1 the Jaccard index value between ($\mathcal{IC}(C_i, C_{i+2})$, $C_{i+1}$), the better the interpolation result is. Table I displays HD values of some contours of both case studies "Zone A", and "Zone B", and Fig.6 displays HD values for $\mathcal{IC}(C_i, C_{i+2})$ and $C_{i+1}$ (as HD interpolated) vs. HD values for $(C_i, C_{i+1})$ (as HD actual) for all the contours with their elevation values. The degree of matching in "Zone A" is seen to be higher than it is for "Zone A" (Fig. 6). In Table I, some of the rows $C_1, C_3, C_{33}, C_{34},$ and $C_{36}$ shows an exact match between Hausdorff Distance values, however, there is exceptions also.

Table I

Hausdorff Distance values for "Zone" A and "Zone B"

| | $C_i$, Elevation (meter) | HD ($\mathcal{IC}(C_i, C_{i+2}), C_{i+1}$) | HD ($C_i, C_{i+1}$) |
|---|---|---|---|
| ZONE A | $C_1, 2340$ | **15** | 15 |
| | $C_2, 2360$ | 7 | 14 |
| | $C_3, 2380$ | **16** | 16 |
| | ⋮ | ⋮ | ⋮ |
| | $C_{49}, 3460$ | 13 | 14 |
| | $C_{50}, 3480$ | 13 | 15 |
| | $C_{51}, 3500$ | 6 | 8 |
| | $C_{52}, 3520$ | 5 | 5 |
| ZONE B | $C_1, 1160$ | **6** | 6 |
| | $C_2, 1180$ | 4 | 6 |
| | $C_3, 1200$ | **6** | 6 |
| | $C_4, 1220$ | 5 | 7 |
| | ⋮ | ⋮ | ⋮ |
| | $C_{33}, 1840$ | **2** | 2 |
| | $C_{34}, 1860$ | **2** | 2 |
| | $C_{35}, 1880$ | 1 | 2 |
| | $C_{36}, 1900$ | **2** | 2 |

The Jaccard Index, also called a similarity coefficient, is used to determine how similar a sample set is to another. We evaluate the similarity between the given contour ($C_{i+1}$) and the interpolated contour ($\mathcal{IC}(C_i, C_{i+2})$) as we consider each contour as a set. The Jaccard index value ranges from 0 to 1, with 0 denoting no similarity and 1 denoting an exact set. More similarity is indicated by a Jaccard similarity coefficient that is closer to 1. Both the case study have Jaccard index values greater than 0.9 (Fig.4 3$^{rd}$ column). It indicates that there is high similarity between the interpolated contours and input contours. However, in case of "Zone B" some of contours are showing a Jaccard Index value of approximately 0.8, indicating slight mismatch with the actual contour set.

Except a few contours for the results of the entire analysis, the overall acceptable range of RMSE, MAPE, and Jaccard Similarity Coefficient and matching of HD values of the actual or predicted contours for both case studies show a high degree of validation of the proposed interpolation method. Some of the discrepancies caused by the high degree of crenulation in the contours. Additionally, some of the contours contain information about abrupt changes in slope, and when we use those contours as test contours, the method is unable to predict those changes.

This entire approach can be extended to other contexts of geosciences that include the generation of high spatial resolution stratigraphic sequences, tree-ring structures, gravity, magnetic, seismological, and resistivity profiles, and contours. This list also includes all those contours such as isotherms, isohyets, etc.

## V. CONCLUSIONS

Interpolation of intermediate contours from existing sparse contour maps is a challenging task in the field of geospatial visualization. Maximum possible recursive intermediate contour interpolation via computing the median elevation region is a way to achieve visualization of the continuous surface from contour map. Our proposed method describes the spatial relationship between the contour regions and gives a simplification of the original problem in terms of sets interpolation based on different categories. The categorization is also done between different TERs and their subsets as source-set and target-set based on the logical relationship. From the result section, it can be seen how the transition takes place from the source elevation region to target elevations and ends up creating an intermediate contour using the mathematical morphological gradient. Results have been shown in synthetic data of contours at the different spatial locations, also in contour maps taken from a real topographic map of Mt. Washington, NH. Further, the method is evaluated quantitatively using RMSE, and MAPE with the test points of 40-meter contour interval contours. To validate how the proposed approach retains the morphological attributes of a contour, we also analyzed by computing the Jaccard Similarity coefficient and Hausdorff Distance between interpolated test contours and given original contours as ground truth data. The proposed framework provides valuable insight into the spatial visualization of discrete contour maps. However, in this article, we only consider this framework on contour maps, assuming that all contours run roughly parallel to each other. As a result, the errors appear to be high in some cases, and because we skipped alternate contours for testing the method, the slope information is also altered. However, the result is an acceptable smooth and artifact-free surface. Future work will investigate how this framework can be used in all other logical relationships and situations involving spatial interpolation.

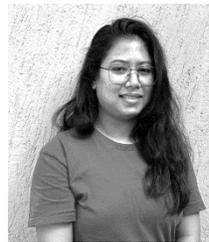

**Geetika Barman** (Member, IEEE) received both a B.Tech degree in Computer Science & Engineering and M.Tech in Information Technology from Tezpur university, Assam, in 20013 and 2017, respectively.

She is currently working as Senior Research fellow at System Science and Informatics Unit, Indian Statistical Institute. Her current research interests includes image processing, remote Sensing image processing using mathematical morphology, hyperspectral image processing. She is a student member of the IEEE.

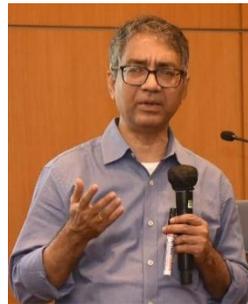

**B. S. DayaSagar** (M'03-SM'03) is a Full Professor (Higher Administrative Grade) of the Systems Science and Informatics Unit (SSIU) at the Indian Statistical Institute. Sagar received his MSc and PhD degrees in Geoengineering and Remote Sensing from the Faculty of Engineering, Andhra University, Visakhapatnam, India, in 1991 and 1994 respectively. He is also first Head of the SSIU. Earlier, he worked in the College of Engineering, Andhra University, and Centre for Remote Imaging Sensing and Processing (CRISP), The National University of Singapore in various positions during 1992-2001. He served as Associate Professor and Researcher in the Faculty of Engineering & Technology (FET), Multimedia University, Malaysia, during 2001-2007. Since 2017, he has been a Visiting Professor at the University of Trento, Trento, Italy. His research interests include mathematical morphology, GISci, digital image processing, fractals and multifractals, their applications in extraction, analyses, and modeling of geophysical patterns. He has published over 90 papers in journals, and has authored and/or guest edited 14 books and/or special theme issues for journals. He recently authored a book entitled "Mathematical

Morphology in Geomorphology and GISci," CRC Press: Boca Raton, 2013, p. 546. He recently co-edited two special issues on "Filtering and Segmentation with Mathematical Morphology" for IEEE Journal of Selected Topics in Signal Processing (v. 6, no. 7, p. 737-886, 2012), and "Applied Earth Observation and Remote Sensing in India" for IEEE Journal of Selected Topics in Applied Earth Observation and Remote Sensing (v. 10, no. 12, p. 5149-5328, 2017). His recent book "Handbook of Mathematical Geosciences", Springer Publishers, p. 942, 2018 reached one million downloads. He is an elected Fellow of Royal Geographical Society (1999), Indian Geophysical Union (2011), Indian Academy of Sciences (2022), and was a member of New York Academy of Science during 1995-1996. He received the Dr. Balakrishna Memorial Award from Andhra Pradesh Academy of Sciences in 1995, the Krishnan Gold Medal from Indian Geophysical Union in 2002, and the "Georges Matheron Award-2011 (with Lecturership)" of the International Association for Mathematical Geosciences. He is the Founding Chairman of Bangalore Section IEEE GRSS Chapter. He is an IEEE Geoscience and Remote Sensing Society (GRSS) Distinguished Lecturer (DL) for 2020-2023. He is a Member of the Honors and Recognition Committee of the American Geophysical Union (AGU). He is on the Editorial Boards of Computers & Geosciences, Frontiers: Environmental Informatics, and Mathematical Geosciences. He is also the Editor-In-Chief of the Springer Publishers's Encyclopedia of Mathematical Geosciences. He is a Co-Editor of the CRC Press Book Series on "Signal and Image Processing of Earth Observations".